# Modality-agnostic, patient-specific digital twins modeling temporally varying digestive motion

Jorge Tapias Gomez[1,2], Nishant Nadkarni[2], Lando S. Bosma[3], Jue Jiang[2], Ergys D. Subashi[4], William P. Segars[5], James M. Balter[6], Mert R Sabuncu[1], Neelam Tyagi[2]* and Harini Veeraraghavan[2]*

[1] Computer and Information Science, Cornell University, New York, United States of America
[2] Department of Medical Physics, Memorial Sloan Kettering Cancer Center, New York, United States of America
[3] University Medical Center Utrecht, Utrecht, Netherlands
[4] Department of Radiation Physics, University of Texas MD Anderson Cancer Center, Houston, United States of America
[5] Carl E. Ravin Advanced Imaging Laboratories and Center for Virtual Imaging Trials, Duke University Medical Center, Durham, United States of America
[6] Department of Radiation Oncology, University of Michigan, Ann Arbor, United States of America

*Co-Senior Author

**Abstract**

*Objective*: Clinical implementation of deformable image registration (DIR) requires voxel-based spatial accuracy metrics such as manually identified landmarks, which are challenging to implement for highly mobile gastrointestinal (GI) organs. To address this, patient-specific digital twins modeling temporally varying motion were created to assess the accuracy of DIR methods. *Approach*: A total of 21 motion phases simulating digestive GI motion as 4D image sequences were generated from static 3D patient scans using published analytical GI motion models through a multi-step semi-automated pipeline. Eleven datasets, including six T2-weighted FSE MRI (T2w MRI), two T1-weighted 4D golden-angle stack-of-stars, and three contrast-enhanced computed tomography scans were analyzed. The motion amplitudes of the digital twins were assessed against real patient stomach motion amplitudes extracted from independent 4D MRI datasets using hierarchical motion reconstruction. The generated digital twins were then used to assess six different DIR methods using target registration error, Dice similarity coefficient, and the 95th percentile Hausdorff distance using summary metrics and voxel-level granular visualizations. Finally, for a subset of T2w MRI scans collected from patients treated with MR-guided radiation therapy, dose distributions were warped and accumulated to assess dose warping errors, including evaluations of DIR performance in both low- and high-dose regions for patient-specific error estimation. *Main results*: Our proposed pipeline synthesized digital twins modeling realistic GI motion, achieving mean and maximum motion amplitudes and a mean log Jacobian determinant within 0.8 mm and 0.01, respectively, similar to published real-patient gastric motion data. It also enables the extraction of detailed quantitative DIR performance metrics and supports rigorous validation of dose mapping accuracy prior to clinical implementation. *Significance*: The developed pipeline enables rigorously testing DIR tools for dynamic, anatomically complex regions facilitating granular spatial and dosimetric accuracies.

## 1. Introduction

Deformable image registration (DIR) is a critical component of both computed tomography (CT) and magnetic resonance (MR) image-guided adaptive radiotherapy [1-3]. It enables automated propagation of organ-at-risk (OAR) segmentations across treatment fractions and facilitates accurate estimation of the radiation dose delivered to both OARs and target structures throughout the treatment course. A key obstacle towards clinical implementation of DIR, particularly for luminal gastrointestinal (GI) organs where motion can range from a few millimeters up to 3 cm [4-6], is the lack of methods to assess DIR accuracy at a patient-specific and granular level.

Target registration error (TRE), which measures the displacement errors of known key points as outlined in TG-132 [7], provides a granular and unbiased assessment of DIR accuracy. In contrast, segmentation overlap metrics are influenced by the size of evaluated structures, potentially limiting their sensitivity and interpretability. However, TRE requires manually or semi-automatically placed visually discernible keypoints, which are difficult to identify reliably in GI organs due to their large and often unpredictable motion across and within radiation treatment fractions. Keypoints are typically placed in high-contrast regions with large intensity gradients, areas where DIR algorithms tend to perform well, potentially biasing the evaluation. Manually placed keypoints may not reflect registration accuracy in low-contrast soft-tissue regions, leading to an overestimation of DIR performance. Previous studies have highlighted the importance of using digital phantoms to address this limitation and evaluate DIR methods for dose accumulation [8] and even generated ground truths to assess the performance of DIR methods for prostate cancer [9]. The goal of this study is to develop a patient-specific digital twin (DT) [10-11] framework to serve as known ground truth for evaluating the accuracy of DIR and accumulated dose estimates in the gastrointestinal tract.

Population-level digital phantoms simulating respiratory and GI motions have been previously developed [12,13]. Prior work by Subashi et.al implemented various GI motion patterns including peristalsis, rhythmic segmentation in the stomach, small and large bowel, high amplitude propagating contractions (HAPC) in the large bowel, and tonic contractions in the GI sphincters using published analytical models within an XCAT computational phantom [12]. Prior studies have also focused on extracting individual motions such as breathing versus stomach contractile motions to extract the amplitude and time scale of such motions from real patients [6,14]. While population-based anatomical models are an important tool for advancing research in medical imaging and radiation therapy [15–21], they fall short in representing subtle, patient-specific anatomical and imaging variations that are essential for the rigorous evaluation of both iterative and deep learning (DL) DIR methods.

We, extend the prior work by Subashi et.al [12], adapting the XCAT-based framework to individual patient anatomy. We implement temporally varying GI motion using published amplitude and time-scale data derived from real patient imaging [6]. Our framework aligns with the definition of DTs set forth by the Ecosystem Digital Twins in Health (EDITH) Coordination and Support Action funded by the European Commission [22]. EDITH defines DTs for health as computer simulations, incorporating both knowledge-driven and data-driven models to predict clinically relevant quantities otherwise challenging to experimentally measure.

Our pipeline starts from individual patient imaging data, extracting the various organs using a semi-automated manner, and creating a digitized representation as non-uniform rational B-spline surfaces (NURBS). A motion pattern is then applied to this model to generate a patient-specific sequence of deformations simulating organ motion. We validate these simulated motion patterns by comparing them to real patient motion data from the literature.

Our key contributions include: (a) development of a patient-specific DT framework that simulates temporally varying GI motion and provides known ground truth for DIR evaluation, (b) demonstration of this framework across multiple imaging modality, including contrast-enhanced computed tomography (CECT), T2-weighted fast spin echo MRI (T2w MRI), and T1-weighted 4D golden-angle stack-of-stars MRI (T1w gaSOS), (c) the first framework, to our knowledge, enabling granular, patient-specific assessment of registration and dosimetric accuracy in the GI tract, and (d) a modality and algorithm independent evaluation framework, demonstrated on both iterative and deep learning-based DIR methods.

We envision our approach to be useful for performing day-to-day and patient-specific assessments of a preferred DIR method used in a clinic for the MR-Linac [21-24] enabled adaptive radiation treatment (ART) or assessing quality of treatment planning simulations.

## 2. Materials and Methods

### 2.1 DT Pipeline for GI Motion Simulation

We develop a semi-automated pipeline to generate patient-specific DTs that simulate temporally varying gastrointestinal motion. The pipeline produces 4D image sequences by deforming static 3D patient scans using analytical GI motion models [12] applied to organ-specific surface representations. Twenty-one motion phases were synthesized to match the temporal resolution of stomach contractions reported by Johanssen et.al [14]. The pipeline comprises the following steps (Figure 1):

(a) Organ segmentation: A published AI model called the self-distilled masked image transformer (SMIT) [27], was used to segment the liver, stomach, duodenum, spleen, kidneys, small and large bowel from a given patient image set (MR or CT). Segmentations were manually verified by an expert (Fig 1. Step 1, row 1).

(b) Organ-specific skeleton graph extraction: Morphological thinning of binary 3D segmentation masks is performed to obtain a minimally connected sequence of points by making successive passes of the image and removing the identified border pixel, while preventing a break in connectivity of points spanning the organ [28]. This skeleton may contain branches which require further processing (Fig 1. Step 1, row 2).

(c) Longest medial axis extraction: The starting and ending points were manually selected from each extracted skeleton and a breadth-first search algorithm [29] was used to determine the longest connected path graph between the selected points. This path was retained as the organ centerline while all branching structures were discarded (Fig 1. Step 1, row 3).

(d) Non-uniform rational B-spline (NURBS) surface extraction: For each organ, radial vectors were cast from sampled points along the centerline to identify boundary intersections with the segmentation mask. These boundary points served as control points to define the non-uniform rational B-spline (NURBS) surface to define the organ's geometry. Each cross-section along the centerline defined a sectional curve (Fig 1. Step 2) for the surface. These NURBS surfaces were then used to synthesize GI motion by applying a motion model at the surface points.

(e) Multi-phase motion synthesis: Analytical motion models developed by Subashi et al. [12] were applied to the NURBS surfaces to simulate realistic GI motion patterns using traveling sinusoidal waves. These include high amplitude propagated contractions (HAPCs), tonic contractions, and peristalsis (Fig 1. Step 3).

The overall modeling framework is given by the following equation:

$$P'_{i,j} = P_{i,j} + F(\vec{x_i}, t) \cdot D_s \cdot D_t \cdot \vec{d}$$

Where $P_{i,j}$ represents the control point $j$ at the sectional curve $i$ of a given NURBS surface. $F(\vec{i}, t)$ is a function modeling the non-dispersive component of the wave, $D_s$ and $D_t$ represent functions modeling dispersions of the wave in the spatial coordinate $(\vec{x_i})$ and temporal ($t$) domain respectively and $\vec{d}$ represents a directional vector to apply a consistent expansion/contraction in the radial direction from the center for each sectional curve.

Using this framework, we iterate through each sectional curve of the given NURBS surface and compute the wave and dispersion magnitudes as functions of space and time. In this work, we specifically focused on modeling the peristaltic motion ($F_{PS}(\vec{x_i}, t)$), within the stomach and large bowel. The wave function was modulated via $F_{PS}(\vec{x_i}, t)$ and an exponential dispersion function $D_z(u)$. Of note, the DT framework is flexible and can be adapted to incorporate any GI motion patterns and dispersion functions. The peristaltic wave and exponential dispersion function are given by:

$$F_{PS}(\vec{x_i}, t) = \frac{1}{\sqrt{3}} \cdot A \cdot \sin\left(2\pi \frac{L_i - c \cdot t}{\lambda}\right)$$

$$D_z(u) = e^{-\alpha \cdot u}$$

where A, c, λ, and α are user-defined parameters representing the amplitude, speed, and wavelength of the peristaltic wave, and the attenuation of amplitude in the spatial or temporal domain, respectively. $L_i$ is the position of the sectional curve along the organ's length, and u is a generalized variable that corresponds to either $\vec{x_i}$ or $t$, and z is an index identifying the spatial ($s$) or temporal ($t$) domain.

Applying the motion models to the NURBS surfaces yields a sequence of deformed surfaces, which serve as the basis for computing the corresponding deformation vector fields (DVFs).

(f) Extraction of deformation vector fields (DVFs): From the sequence of deformed NURBS surfaces, we can derive a 3D DVF that captures the deformation for each NURBS by:

**Computing Inner-Shell Vectors Using Radial Interpolation.** To capture internal surface deformation within each GI organ, intermediate "inner shell" surfaces were generated between the NURBS outer surface and the organ's centerline. These shells were parameterized by a radial factor $p \in [0,1]$, with $p = 0$ at the outer surface and $p = 1$ at the centerline. For each shell defined by $p_k$, a new NURBS surface was computed by interpolating between the outer and centerline surfaces. Sampling each shell over a uniform $(u, v)$ grid yielded corresponding 3D points in both the original and deformed configurations. These parameters are standard in NURBS surface definitions, where u and v define positions within the 2D parameter space of the surface.

**Deformation Vectors:** Deformation vectors were then calculated pointwise as the difference between the deformed and original positions: $V_{surf}(u_i\,v_j, p_k) = X'(u_i\,v_j, p_k) - X(u_i\,v_j, p_k)$

**Voxelization and Smoothing of the DVF:** The computed deformation vectors across all $(u, v, p)$ samples were mapped into a fixed 3D voxel grid by assigning each vector to a voxel based on its original undeformed position. When multiple vectors mapped into the same voxel, they were averaged. To improve continuity, a smoothing step was applied wherein zero-motion voxels were iteratively updated by averaging the vectors of neighboring non-zero voxels.

This will generate a DVF that can be applied to the patient's static scan to deform it in a way that reflects both surface and internal anatomical displacements (Fig. 1, Step 4).

(g) <u>Creation of 4D MRI/ 4D CT sequences:</u> By repeating the previous step across the full sequence of deformed NURBS surfaces, we synthesize patient-specific motion that is agnostic to any imaging modality. Applying the corresponding DVFs to the original patient scans produces a 4D sequence of GI motion while preserving the internal contents of the luminal organs (e.g. air, fluid). Sample videos of synthetized motion sequences are included in supplemental material for representative cases.

### 2.2 Datasets and Motion Parameters

Under institutional review board-aproved protocols, a total of 11 datasets were analyzed, comprising six T2w MRI, two T1w gaSOS, and three CECT scans. Additionally, three independent T1w gaSOS datasets were used solely for quality assurance, with motion models tailored to the motion patterns observed in those specific cases. To the stomach, we applied a traveling wave equation with parameters A=16 mm, λ=55 mm, and c=5 mm/sec. In addition, for the five datasets that included the large bowel, a second traveling wave equation A=16 mm, λ=40 mm, c=8 mm/sec was applied to simulate large bowel motion. For all synthesized motions used in the DIR evaluation, α=0 was chosen to emulate high-magnitude, non-dispersive GI motion.

### 2.3 Patient-Specific Motion Quality Assurance

To evaluate motion realism, we compared synthetic stomach motion from the DT framework to motion derived from three T1w gaSOS MRI datasets with known stomach contractions. Specifically, we used the previously published known stomach deformation motions, which includes detailed recordings of stomach contractions over 21 distinct phases. Previous studies have used hierarchical MRI reconstructions to model stomach contractions and slow drifts with and without respiratory motion [6]. Employing the ground truth DVFs derived from three scans with hierarchical motion models as baseline references, we fit traveling sinusoidal waves tailored specifically to each contraction phase. This allowed the DT pipeline to generate synthetic stomach motions closely aligned with the experimentally measured motions. To validate the synthesized motion, we computed the mean and maximum displacements of the ground-truth DVFs and the log of the Jacobian which captures the degree of local non-rigid, non-affine deformations.

### 2.4 Motion Analysis

For all 21 synthetically generated motion phases across the 11 patient scans, the maximum and mean displacement magnitudes were computed for the full body as well as two gastric organs: the stomach and large bowel.

### 2.5 Evaluation of DIR Methods using DTs

The goal of the experiment was to assess the feasibility of using the DTs to evaluate a range of DIR methods using the synthesized GI motions produced with various imaging modalities for multiple patients. Of note, the goal was not to compare individual methods with respect to one another; all methods were used as is without performing any hyperparameter or other optimization for the evaluated DT datasets.

Six different DIR algorithms were assessed including 4 variational and 2 DL methods. The variational methods included Horn-Schunck optical flow (HSOF), EVolution multimodality (EVO), Elastix with mutual information, and the Iterative Demons algorithm, all of which used iterative optimization using image intensities. Key differences between the methods include the use of brightness consistency and smoothness regularization for motion fields used in HSOF [30], normalized intensity gradients with smoothness regularization used in EVO [31], B-spline parametrization of the transformation used in Elastix [32, 33] and optical flow with Gaussian smoothing in diffeomorphic Demons [34, 35]. The two DL DIR methods were VoxelMorph [36] and an enhancement of Voxelmorph called progressively refined registration-segmentation (ProRSeg) that uses convolutional long short-term memory networks in the encoder [37]. Testing datasets used for evaluation were never used for training VoxelMorph and ProRSeg to avoid data leaks. Both VoxelMorph and ProRSeg were trained on different sets of real patient T2w datasets used in [37] and were only applied to DTs corresponding to the same MR scanning sequence.

### 2.6 Evaluation Metrics

Registration was performed between the original scan and the phase exhibiting the maximum deformation. Geometric accuracy was computed by measuring organ segmentation accuracy using the Dice similarity coefficient (DSC) and Hausdorff distance at 95th percentile as well as Target Registration Error (TRE) using the surface and inner shell points of a given NURBS surface. Differences in dose accumulation across methods were assessed by simulating dose accumulation using both the ground truth DVF and the DVFs produced by each registration method. Dose deformation was performed using direct dose mapping. Dose warping error (DWE) comparing the accumulated dose distributions was calculated as:

$$DWE = \sum_{i=1}^{N} \frac{Accum\ D_{i,DIR} - Accum\ D_{i,GT}}{Accum\ D_{i,GT}}$$

Granular assessment of dosimetric and displacement errors were computed at the anatomic voxel level and quantified using Root Mean Squared Error (RMSE), which then were also visualized as heatmaps for a more detailed evaluation of model accuracy in regions subject to large deformations or high-radiation dose exposure.

## 3. Results

Our semi-automated pipeline was applied to generate motion for 11 different patients from various anatomic imaging modalities. On average, the framework took 45 mins to synthesize motion starting from a 3D input image. The framework was capable of generating motion for image volumes and voxels with varying with varying sizes, as shown in Table 1. The method was applicable to axial and coronal image reconstructions.

### 3.1 Assessing Patient-Specific Motion Quality Assurance

The top row of Figure 2 shows a comparison between DT-generated stomach motion and known stomach motion derived from Zhang et al. using hierarchical motion modeling [6], showing both mean and maximum displacement magnitudes. Across all motion phases, the DT-generated motion exhibited mean and maximum displacement differences within 0.8 mm of ground truth. Furthermore, as shown in the bottom row of Figure 2, the DT-simulated deformations yielded consistent log-Jacobian means across all motion phases, remaining within 0.01 of the ground truth mean over all 21 phases. This indicates strong agreement in the overall deformation characteristics.

This quality assurance analysis indicates that the synthesized stomach motions can closely match a known reference values, both in displacement magnitude and deformation characteristics.

### 3.2 Modality Agnostic Temporally Varying Motion Synthesis

Figure 3. depicts the synthesized motions with the corresponding DVFs for all three analyzed imaging modalities, CECT, T1w gaSOS, and T2w MRI. The deformed masks and 2D image slices taken at different time points along the traveling wave sequence propagating through the stomach are also shown. The maximum stomach motion shown in Table 2 ranged from 8.56 mm to 14.34 mm, while the large bowel exhibited motion magnitudes between 7.69 mm and 8.64 mm, indicating the ability to simulate motion at varying amplitudes. The mean and standard deviation in the motion magnitudes for the gastric organs and the whole body are summarized in Table 3, which showed stomach mean displacements ranging from 0.92 mm to 3.69 mm, and the large bowel from 0.77 mm to 2.96 mm.

This analysis shows that our framework can generate realistic motion for a variety of organs with differing geometries across multiple common radiological imaging modalities, producing motion magnitudes consistent with those reported by Zhang et al [6].

### 3.3 Evaluating DIR Performance using the DT Framework

DIR was computed using various methods between the phase 0 (static input patient 3D scan) and the phase with the largest amplitude displacement with respect to phase 0. Variational DIR methods were applied to the CECT, T1w gaSOS, and T2w MRI, whereas the DL-DIRs trained with T2w MRI were only applied to T2w MRI datasets.

Table 4 shows the accuracy metrics computed for the variational methods applied to the CECT and T1w gaSOS scans. The DIR methods were similarly accurate for T1w gaSOS compared to CECT using TRE ($2.14 \pm 1.31$ mm to $3.64 \pm 2.12$ mm versus $2.48 \pm 1.33$ mm to $3.60 \pm 2.00$ mm), HD95 (6.00 mm to 8.77mm versus 1 mm to 8.83 mm), and DSC (0.68 to 0.81 versus 0.71 to 0.99). HD95 and DSC were computed for two CECT patients and three T1w gaSOS cases; however, standard deviation values are not reported due to the limited sample size.

In the case of T2w MRI from 5 patients with pancreatic cancer, the DIR methods showed slightly higher TRE $4.14 \pm 2.23$ mm to $5.14 \pm 2.51$ mm for the stomach and duodenum compared to T1WI gaSOS and CECT images, a higher DSC from $0.84 \pm 0.02$ to $0.92 \pm 0.02$, and lower HD95 ranging from $2.71 \pm 0.51$ mm to $4.4 \pm 0.18$ mm. The registration accuracies for the large bowel were slightly lower than the stomach and duodenum with TRE ranging from $3.06 \pm 0.48$ mm to $4.35 \pm 0.29$ mm, DSC between $0.89 \pm 0.02$ to $0.95 \pm 0.01$, and HD95 between $1.94 \pm 0.65$ mm and $3.82 \pm 0.44$ mm, respectively.

In addition, the same 5 patients also had radiation treatment dose maps, which were used to calculate the DWE for the same three organs. As shown in Figure 4, the mean DWE ranged from $6.68 \pm 1.45$ % to $9.80 \pm 1.83$ % for stomach and duodenum and a smaller error of $3.88 \pm 1.00$ % to $6.72 \pm 1.81$ % for the large bowel. The results also show variation in the performance of the different methods across the different datasets and modalities, thus providing an approach to assess the relative merits of the various methods for segmentation, registration, and dose warping on an individual patient level.

### 3.4 Patient-Specific Granular Errors

Capability to assess granular (or voxel-level) error visualization on a patient level is demonstrated for two representative patients, patient A (Figure 5) where organs undergoing motion occurred in the low-radiation dose regions (10 to 30 Gy) (see figure 5d) and patient B (Figure 6) with the same organs located in the high radiation dose regions (exceeding 30 Gy) (see figure 6d). Global root mean square error (RMSE) is visualized with respect to increasing motion magnitudes to assess errors as a function of motion (Figure 5a, Figure 6a). Motion magnitude was binned at fixed intervals ranging from a minimum of 0 mm to a maximum of 8.65 mm. Patient specific analysis showed that the mean RMSE ranged from 0.7 mm in low-motion regions (0–1 mm) to 5 mm in high-motion regions (>8 mm) for the two patients. RMSE was also computed within the GI organs undergoing motion to assess impact of motion on the accuracy with respect to the radiation dose delivered to the organ, which showed a range from 0.10 mm to 2.00 mm, with the highest errors occurring mostly in low-dose regions for Patient A, but a higher error of 1.00 mm to 2.5 mm in the high-dose region for Patient B (Figure 5b, Figure 6b).

Figure 5c, 6c and Figure 5d, 6d show a visualization of voxel-wise RMSE within the organs undergoing motion and the dose maps, respectively to provide a visual representation of the errors for individual patients.

## 4. Discussion

In this work, we extended the concept of population-level digital phantoms to develop patient-specific DTs that model temporally varying gastrointestinal motion. Our semi-automated pipeline starts from AI automated organ segmentations, which then are used to generate peristaltic motion of varying amplitudes and time scales for luminal organs such as the stomach, duodenum and the large bowel. Our framework is applicable to multiple anatomic imaging modalities and demonstrated feasibility to evaluate multiple variational and two different deep learning DIR methods. In addition to evaluating registration, our framework can be easily extended to evaluate dose warping accuracy summarizing errors for individual organs as well as on a granular level to assess accuracy variations on voxel-level. As a result, our DT framework enables voxel-wise visualization of registration errors, facilitates analysis of error patterns across motion regimes (e.g. low vs high motion), and supports individualized assessment of how registration inaccuracies can affect radiation dose. To our knowledge, this is the first comprehensive simulation and evaluation of DIR using patient-specific DTs across multiple imaging modalities for GI luminal organs.

Whereas, previous efforts focused on modeling respiratory and cardiac motion, ours focused on modeling the digestive motion [13, 15, 39, 40]. One prior work by Subashi et.al demonstrated the ability to generate MR-like digital phantoms incorporating a range of GI motion types—including peristalsis, slow and fast gastric contractions, and high-amplitude propagating contractions (HAPCs) [12]. However, all aforementioned prior works synthesized digital phantoms modeling population-level anatomy built on the generic adult male and female XCAT models. A limitation with modeling population-level motion is that it doesn't represent individual patient anatomy variations. Our work, for the first time, addresses the key issue of modeling patient-specific variations by creating patient-specific digital twins of gastric motion.

Another limitation of population level modeling using XCAT requires simulation of MR images using fixed signal intensities for each organ, that can create a domain shift for assessing deep learning registration methods. In our work, synthesis starts from the original MRI, the synthesized motions are also created on MRI, hence allowing to evaluate DL DIR methods.

Finally, GI organs undergo substantial and arbitrary motion that varies from patient to patient despite common motion mitigation strategies such as pneumatic compression belts [4-6, 38]. Our approach allows to vary and create a variety of motion amplitudes and rigorously evaluate DIR methods under various GI motion amplitudes.

Limitations of our current framework include the lack of support for modeling respiratory motion as the focus of this work was isolated GI motion. We also excluded small bowel motion simulation because it is generally segmented in the clinic as a "bowel-bag", making the extraction of NURBS to model motion along the tubular region difficult. Nevertheless, our approach is not limited to GI organs and motion simulation beyond peristaltic motion can be performed for organs such as the liver and kidneys and extension to organs in the pelvis such as the rectum and bladder could be performed using the same framework.

## 5. Conclusion

We developed a semi-automated DT pipeline to generate realistic GI temporally varying motion in the stomach, duodenum and large bowel from multiple anatomic imaging modalities. Our framework showed capability to generate motions within ranges seen in real patients, indicating feasibility to evaluate multiple DIR methods. Our framework enables evaluating dose warping and registration errors in a granular voxel-wise manner for individualized patient-level analysis, suitable for rigorous analysis required for clinical deployment.

## Acknowledgements

This research was supported by NIBIB R01EB032825 and partially supported by the NIH/NCI Cancer Center Support Grant/Core Grant (P30 CA008748)

## References

[1] Meyer S, Hu YC, Rimner A, Mechalakos J, Cerviño L, Zhang P. Deformable Image Registration Uncertainty-Encompassing Dose Accumulation for Adaptive Radiation Therapy. Int J Radiat Oncol Biol Phys. 2025 Apr 14:S0360-3016(25)00371-2. doi: 10.1016/j.ijrobp.2025.04.004. Epub ahead of print. PMID: 40239820.

[2] Lowther N, Louwe R, Yuen J, Hardcastle N, Yeo A, Jameson M; Medical Image and Registration Special Interest Group (MIRSIG) of the ACPSEM. MIRSIG position paper: the use of image registration and fusion algorithms in radiotherapy. Phys Eng Sci Med. 2022 Jun;45(2):421-428. doi: 10.1007/s13246-022-01125-3. Epub 2022 May 6. PMID: 35522369; PMCID: PMC9239966.

[3] Rigaud B, Simon A, Castelli J, Lafond C, Acosta O, Haigron P, Cazoulat G, de Crevoisier R. Deformable image registration for radiation therapy: principle, methods, applications and evaluation. Acta Oncol. 2019 Sep;58(9):1225-1237. doi: 10.1080/0284186X.2019.1620331. Epub 2019 Jun 3. PMID: 31155990.

[4] Mostafaei F, Tai A, Omari E, Song Y, Christian J, Paulson E, Hall W, Erickson B, Li XA. Variations of MRI-assessed peristaltic motions during radiation therapy. PLoS One. 2018 Oct 25;13(10):e0205917. doi: 10.1371/journal.pone.0205917. PMID: 30359413; PMCID: PMC6201905.

[5] Liu L, Johansson A, Cao Y, Kashani R, Lawrence TS, Balter JM. Modeling intra-fractional abdominal configuration changes using breathing motion-corrected radial MRI. Phys Med Biol. 2021 Apr 12;66(8):10.1088/1361-6560/abef42. doi: 10.1088/1361-6560/abef42. PMID: 33725676; PMCID: PMC8159899.

[6] Zhang Y, Kashani R, Cao Y, Lawrence TS, Johansson A, Balter JM. A hierarchical model of abdominal configuration changes extracted from golden angle radial magnetic resonance imaging. Phys Med Biol. 2021 Feb 9;66(4):045018. doi: 10.1088/1361-6560/abd66e. PMID: 33361579; PMCID: PMC7993537.

[7] Brock, K.K., Mutic, S., McNutt, T.R., Li, H. and Kessler, M.L. (2017), Use of image registration and fusion algorithms and techniques in radiotherapy: Report of the AAPM Radiation Therapy Committee Task Group No. 132. Med. Phys., 44: e43-e76. https://doi.org/10.1002/mp.12256

[8] Papachristou K, Katsakiori PF, Papadimitroulas P, Strigari L, Kagadis GC. Digital Twins' Advancements and Applications in Healthcare, Towards Precision Medicine. J Pers Med. 2024 Nov 11;14(11):1101. doi: 10.3390/jpm14111101. PMID: 39590593; PMCID: PMC11595921.

[9] Katsoulakis E, Wang Q, Wu H, Shahriyari L, Fletcher R, Liu J, Achenie L, Liu H, Jackson P, Xiao Y, Syeda-Mahmood T, Tuli R, Deng J. Digital twins for health: a scoping review. NPJ Digit Med. 2024 Mar 22;7(1):77. doi: 10.1038/s41746-024-01073-0. PMID: 38519626; PMCID: PMC10960047.

[10] Lando S. Bosma, Mohammad Hussein, Michael G. Jameson, Soban Asghar, Kristy K. Brock, Jamie R. McClelland, Sara Poeta, Johnson Yuen, Cornel Zachiu, Adam U. Yeo, Tools and recommendations for commissioning and quality assurance of deformable image registration in radiotherapy, Physics and Imaging in Radiation Oncology, https://doi.org/10.1016/j.phro.2024.100647.

[11] Bosma LS, Zachiu C, Ries M, Denis de Senneville B, Raaymakers BW. Quantitative investigation of dose accumulation errors from intra-fraction motion in MRgRT for prostate cancer. Phys Med Biol. 2021 Mar 2;66(6):065002. doi: 10.1088/1361-6560/abe02a. PMID: 33498036.

[12] Subashi E, Segars P, Veeraraghavan H, Deasy J, Tyagi N. A model for gastrointestinal tract motility in a 4D imaging phantom of human anatomy. *Med. Phys.* 2023; 50: 3066–3075. https://doi.org/10.1002/mp.16305

[13] Segars WP, Bond J, Frush J, Hon S, Eckersley C, Williams CH, Feng J, Tward DJ, Ratnanather JT, Miller MI, Frush D, Samei E. Population of anatomically variable 4D XCAT adult phantoms for imaging research and optimization. Med Phys. 2013 Apr;40(4):043701. doi: 10.1118/1.4794178. PMID: 23556927; PMCID: PMC3612121.

[14] Johansson A, Balter JM, Cao Y. Gastrointestinal 4D MRI with respiratory motion correction. Med Phys. 2021 May;48(5):2521-2527. doi: 10.1002/mp.14786. Epub 2021 Mar 24. PMID: 33595909; PMCID: PMC8172093.

[15] Segars WP, Lalush DS, Tsui BM. A realistic spline-based dynamic heart phantom. IEEE Trans. Nucl. Sci. 1999; 46(3): 503-506.

[16] Caon M. Voxel-based computational models of real human anatomy: a review. Radiat Environ Biophys. 2004 Feb;42(4):229-35. doi: 10.1007/s00411-003-0221-8. Epub 2004 Jan 17. PMID: 14730450.

[17] W. P. Segars, D. S. Lalush and B. M. W. Tsui, "A realistic spline-based dynamic heart phantom," 1998 IEEE Nuclear Science Symposium Conference Record. 1998 IEEE Nuclear Science Symposium and Medical Imaging Conference (Cat. No.98CH36255), Toronto, ON, Canada, 1998, pp. 1175-1178 vol.2, doi: 10.1109/NSSMIC.1998.774369.

[18] Bosca RJ, Jackson EF. Creating an anthropomorphic digital MR phantom--an extensible tool for comparing and evaluating quantitative imaging algorithms. Phys Med Biol. 2016 Jan 21;61(2):974-82. doi: 10.1088/0031-9155/61/2/974. Epub 2016 Jan 7. PMID: 26738776.

[19] Ding A, Mille MM, Liu T, Caracappa PF, Xu XG. Extension of RPI-adult male and female computational phantoms to obese patients and a Monte Carlo study of the effect on CT imaging dose. Phys Med Biol. 2012 May 7;57(9):2441-59. doi: 10.1088/0031-9155/57/9/2441. Epub 2012 Apr 5. PMID: 22481470; PMCID: PMC3329718.

[20] P. B. Johnson, S. R. Whalen, M. Wayson, B. Juneja, C. Lee and W. E. Bolch, "Hybrid Patient-Dependent Phantoms Covering Statistical Distributions of Body Morphometry in the U.S. Adult and Pediatric Population," in *Proceedings of the IEEE*, vol. 97, no. 12, pp. 2060-2075, Dec. 2009, doi: 10.1109/JPROC.2009.2032855.

[21] Cassola VF, Milian FM, Kramer R, de Oliveira Lira CA, Khoury HJ. Standing adult human phantoms based on 10th, 50th and 90th mass and height percentiles of male and female Caucasian populations. Phys Med Biol. 2011 Jul 7;56(13):3749-72. doi: 10.1088/0031-9155/56/13/002. Epub 2011 May 31. PMID: 21628776.

[22] EDITH: European Virtual Human Twin . 2022. Accessed November 1, 2023. https://www.edith-csa.eu/

[23] Lagendijk JJ, Raaymakers BW, van Vulpen M. The magnetic resonance imaging-linac system. Semin Radiat Oncol.

[23] 2014 Jul;24(3):207-9. doi: 10.1016/j.semradonc.2014.02.009. PMID: 24931095.

[24] Mutic S, Dempsey JF. The ViewRay system: magnetic resonance-guided and controlled radiotherapy. Semin Radiat Oncol. 2014 Jul;24(3):196-9. doi: 10.1016/j.semradonc.2014.02.008. PMID: 24931092.

[25] Keall PJ, Barton M, Crozier S; Australian MRI-Linac Program, including contributors from Ingham Institute, Illawarra Cancer Care Centre, Liverpool Hospital, Stanford University, Universities of Newcastle, Queensland, Sydney, Western Sydney, and Wollongong. The Australian magnetic resonance imaging-linac program. Semin Radiat Oncol. 2014 Jul;24(3):203-6. doi: 10.1016/j.semradonc.2014.02.015. PMID: 24931094.

[26] Fallone BG. The rotating biplanar linac-magnetic resonance imaging system. Semin Radiat Oncol. 2014 Jul;24(3):200-2. doi: 10.1016/j.semradonc.2014.02.011. PMID: 24931093.

[27] Jiang J, Tyagi N, Tringale K, Crane C, Veeraraghavan H. Self-supervised 3D anatomy segmentation using self-distilled masked image transformer (SMIT). Med Image Comput Comput Assist Interv. 2022 Sep;13434:556-566. doi: 10.1007/978-3-031-16440-8_53. Epub 2022 Sep 16. PMID: 36468915; PMCID: PMC9714226.

[28] T.-C. Lee, R.L. Kashyap and C.-N. Chu, Building skeleton models via 3-D medial surface/axis thinning algorithms. Computer Vision, Graphics, and Image Processing, 56(6):462-478, 1994.

[29] C. Y. Lee, "An Algorithm for Path Connections and Its Applications," in *IRE Transactions on Electronic Computers*, vol. EC-10, no. 3, pp. 346-365, Sept. 1961, doi: 10.1109/TEC.1961.5219222.

[30] Horn, Berthold & Schunck, Brian. (1981). Determining Optical Flow. Artificial Intelligence. 17. 185-203. 10.1016/0004-3702(81)90024-2.

[31] Senneville, Baudouin & Zachiu, Cornel & Ries, Mario & Moonen, C. (2016). EVolution: an Edge-based Variational method for non-rigid multi-modal image registration. Physics in Medicine and Biology. 61. 10.1088/0031-9155/61/20/7377.

[32] S. Klein, M. Staring, K. Murphy, M. A. Viergever and J. P. W. Pluim, "elastix: A Toolbox for Intensity-Based Medical Image Registration," in *IEEE Transactions on Medical Imaging*, vol. 29, no. 1, pp. 196-205, Jan. 2010, doi: 10.1109/TMI.2009.2035616

[33] Leibfarth, S., Mönnich, D., Welz, S., Siegel, C., Schwenzer, N., Schmidt, H., … Thorwarth, D. (2013). A strategy for multimodal deformable image registration to integrate PET/MR into radiotherapy treatment planning. *Acta Oncologica*, *52*(7), 1353–1359. https://doi.org/10.3109/0284186X.2013.813964

[34] Thirion JP. Image matching as a diffusion process: an analogy with Maxwell's demons. Med Image Anal. 1998 Sep;2(3):243-60. doi: 10.1016/s1361-8415(98)80022-4. PMID: 9873902.

[35] Vercauteren T, Pennec X, Perchant A, Ayache N. Diffeomorphic demons: efficient non-parametric image registration. Neuroimage. 2009 Mar;45(1 Suppl):S61-72. doi: 10.1016/j.neuroimage.2008.10.040. Epub 2008 Nov 7. PMID: 19041946.

[36] G. Balakrishnan, A. Zhao, M. R. Sabuncu, J. Guttag and A. V. Dalca, "VoxelMorph: A Learning Framework for Deformable Medical Image Registration," in IEEE Transactions on Medical Imaging, vol. 38, no. 8, pp. 1788-1800, Aug. 2019, doi: 10.1109/TMI.2019.2897538.

[37] Jiang J, Hong J, Tringale K, et al. Progressively refined deep joint registration segmentation (ProRSeg) of gastrointestinal organs at risk: Application to MRI and cone-beam CT. *Med Phys*. 2023; 50: 4758–4774. https://doi.org/10.1002/mp.16527

[38] Alam S, Veeraraghavan H, Tringale K, Amoateng E, Subashi E, Wu AJ, Crane CH, Tyagi N. Inter- and intrafraction motion assessment and accumulated dose quantification of upper gastrointestinal organs during magnetic resonance-guided ablative radiation therapy of pancreas patients. Phys Imaging Radiat Oncol. 2022 Feb 17;21:54-61. doi: 10.1016/j.phro.2022.02.007. PMID: 35243032; PMCID: PMC8861831.

[39] Segars WP, Veress AI, Sturgeon GM, Samei E. Incorporation of the Living Heart Model into the 4D XCAT Phantom for Cardiac Imaging Research. IEEE Trans Radiat Plasma Med Sci. 2019 Jan;3(1):54-60. doi: 10.1109/TRPMS.2018.2823060. Epub 2018 Apr 4. PMID: 30766954; PMCID: PMC6370029.

[40] Segars WP, Tsui BMW, Jing Cai, Fang-Fang Yin, Fung GSK, Samei E. Application of the 4-D XCAT Phantoms in Biomedical Imaging and Beyond. IEEE Trans Med Imaging. 2018 Mar;37(3):680-692. doi: 10.1109/TMI.2017.2738448. Epub 2017 Aug 10. PMID: 28809677; PMCID: PMC5809240.

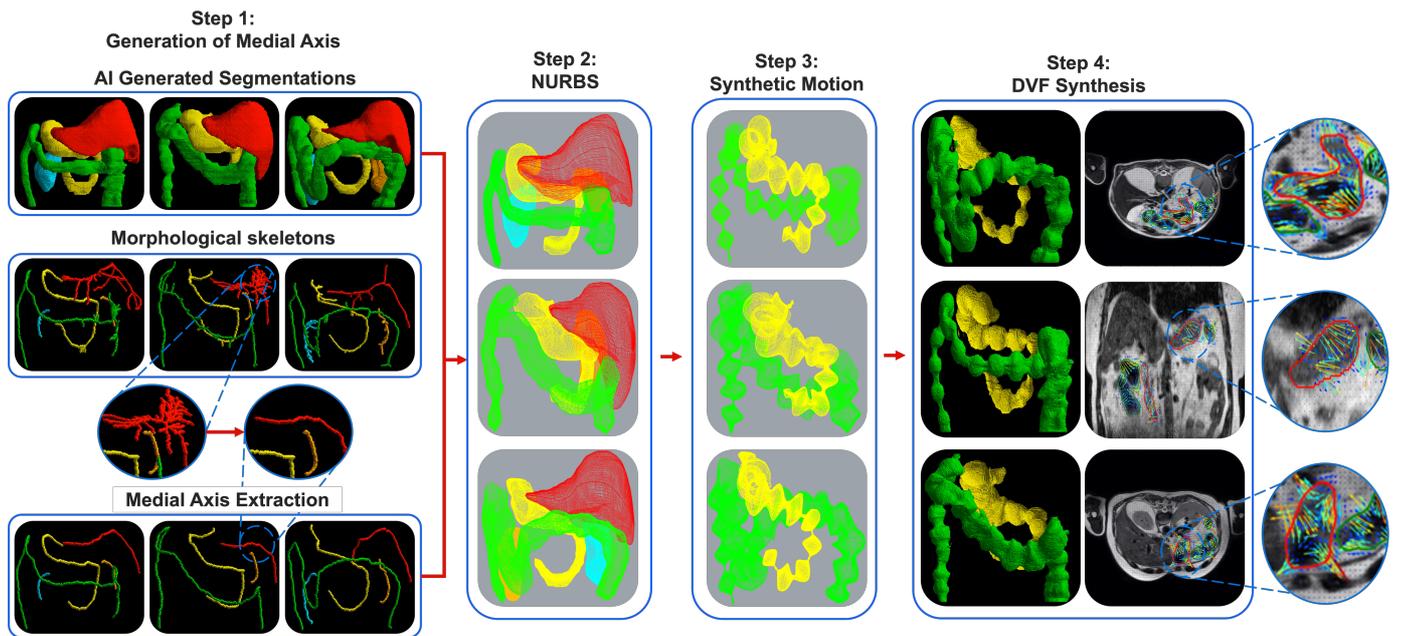

*Figure 1 Overview of the pipeline used to generate a patient-specific DT from a 3D abdominal scan. Step 1: Extract the medial axis by skeletonizing and pruning the AI-generated segmentation masks. Step 2: Generate a NURBS surface based on the medial axis. Step 3: Apply peristaltic motion to the target organs (stomach and large bowel), resulting in 21 phases representing different contraction states. Step 4: Compute the DVFs using the original and deformed NURBS surfaces.*

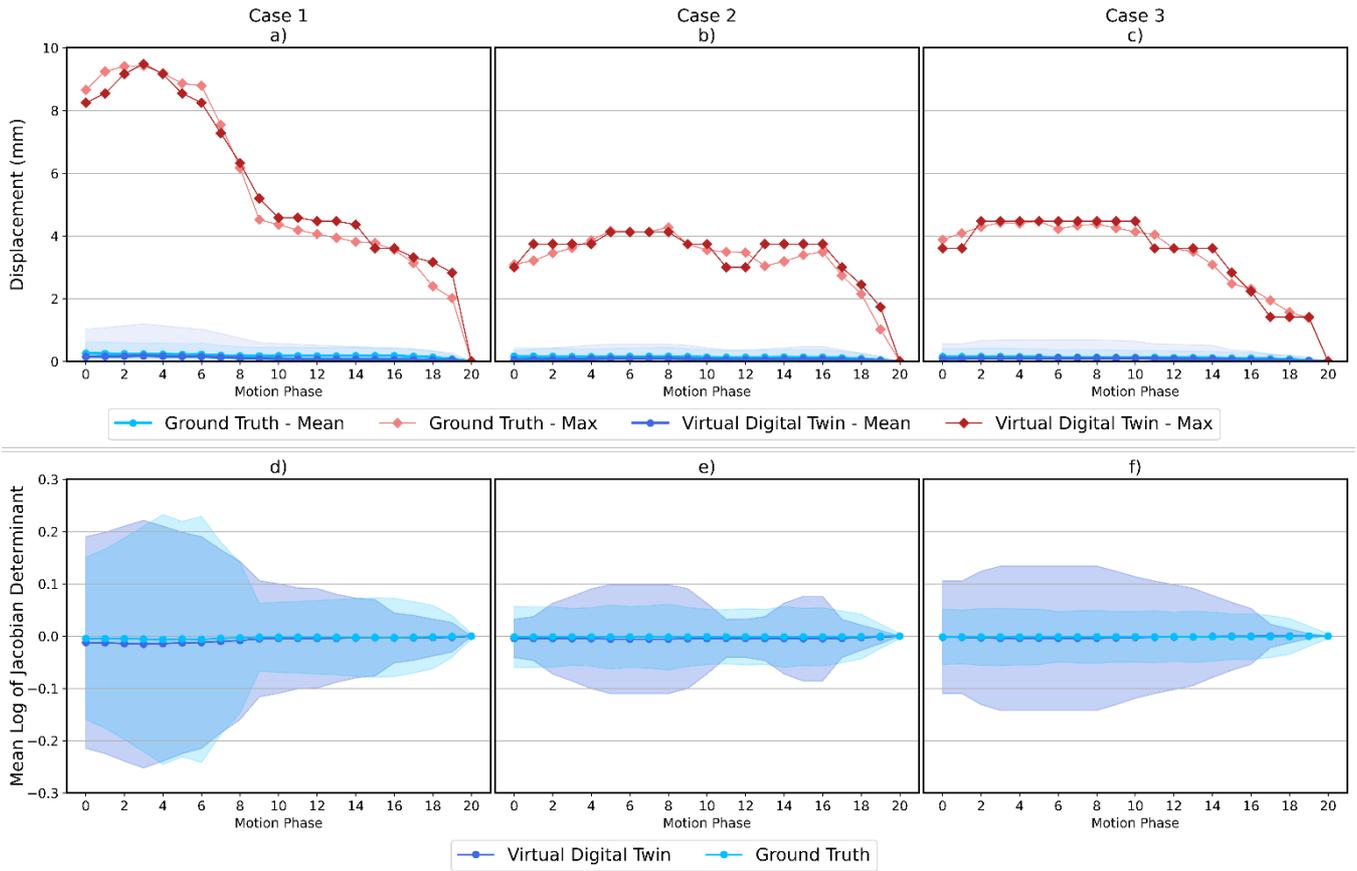

*Figure 2. Top plots (a,b,c): Comparison of mean and maximum displacement magnitudes (± standard deviation) between ground truth and DT-synthetic stomach deformations across 21 phases in three distinct 4D-MRI datasets. Bottom plots (d,e,f): Corresponding comparison of the mean log-Jacobian determinant (± standard deviation), reflecting local volumetric changes during deformation. Together, these plots illustrate both the extent (top) and anatomical plausibility (bottom) of the predicted motion fields relative to ground truth.*

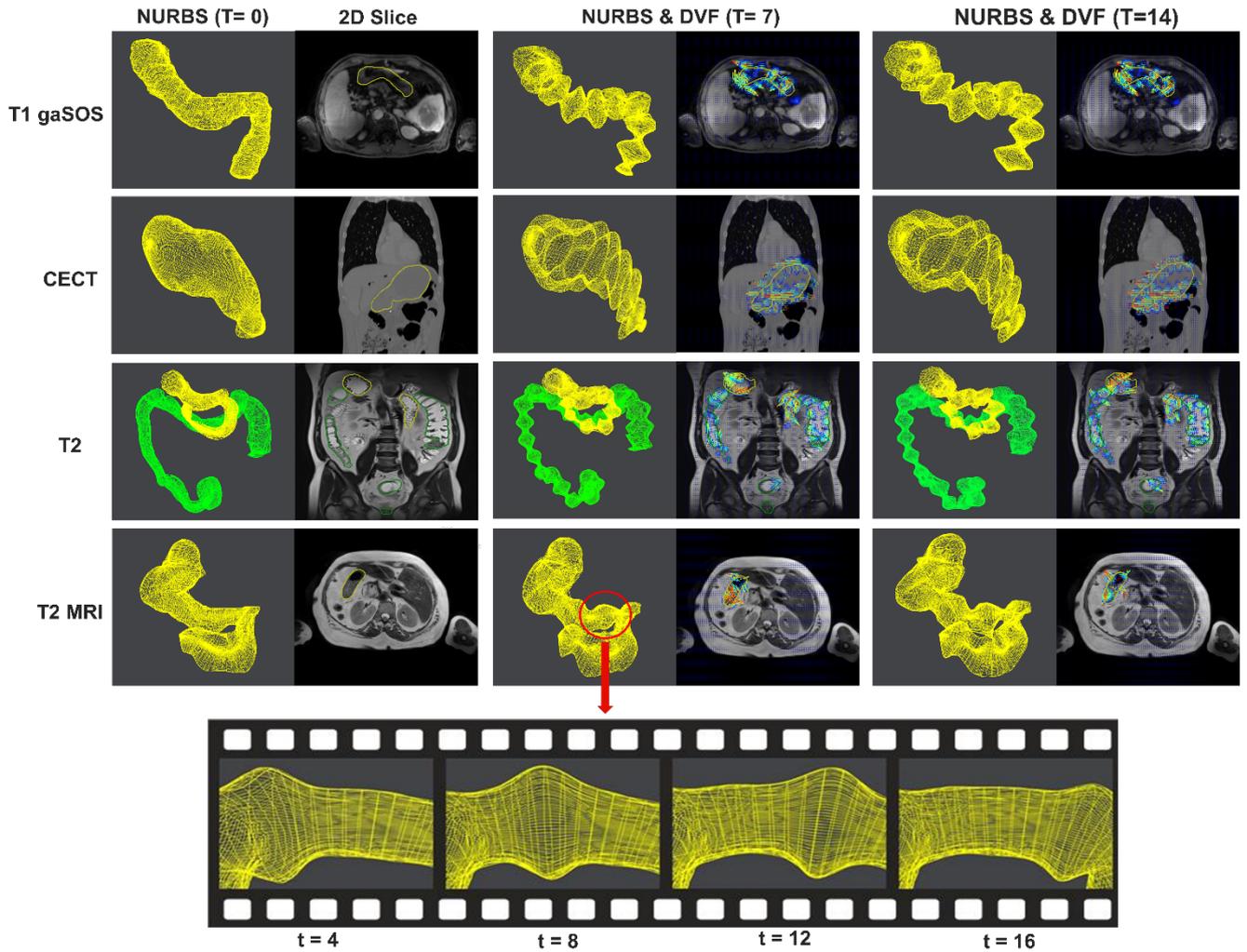

*Figure 3 Example deformed segmentation masks and corresponding DVFs shown for each of the four imaging modalities: CECT, T1w gaSOS, T2w MRI for 3 different time points. Additionally, for the T2w MRI we show a sequence of snapshots of a stomach NURBS surface over four motion phases, illustrating the progression of the applied wave-like motion traversing the organ. The sectional views highlight how the modeled peristaltic deformation propagates along the surface.*

|  | MR Sequence Details | Volume Size | Voxel Size (mm) |
|---|---|---|---|
| Patient 1 | T1w gaSOS | 224 x 224 x 96 | 2.188 x 2.188 x 3.5 |
| Patient 2 | T1w gaSOS | 224 x 224 x 96 | 2.188 x 2.188 x 3.5 |
| Patient 3 | CECT | 512 x 512 x 201 | 0.9766 x 0.9766 x 2 |
| Patient 4 | CECT | 512 x 512 x 201 | 0.9766 x 0.9766 x 2 |
| Patient 5 | CECT | 512 x 512 x 201 | 0.9766 x 0.9766 x 2 |
| Patient 6 | T2w MRI | 512 x 512 x 50 | 0.7813 x 0.7813 x 5 |
| Patient 7 | T2w MRI | 448 x 448 x 125 | 1 x 1 x 2 |
| Patient 8 | T2w MRI | 448 x 448 x 125 | 1 x 1 x 2 |
| Patient 9 | T2w MRI | 448 x 448 x 125 | 1 x 1 x 2 |
| Patient 10 | T2w MRI | 448 x 448 x 125 | 1 x 1 x 2 |
| Patient 11 | T2w MRI | 448 x 448 x 125 | 1 x 1 x 2 |

*Table 1 Details of the image volumes and voxel dimensions used for motion synthesis across 11 patient datasets.*

'

|  | **Full Body** (mm) | **Stomach** (mm) | **Large Bowel** (mm) |
| --- | --- | --- | --- |
| Patient 1 | 13.17 | 13.17 | - |
| Patient 2 | 14.16 | 13.92 | - |
| Patient 3 | 13.22 | 13.22 | - |
| Patient 4 | 13.52 | 13.06 | - |
| Patient 5 | 13.51 | 13.42 | - |
| Patient 6 | 14.34 | 14.34 | - |
| Patient 7 | 8.68 | 8.65 | 8.57 |
| Patient 8 | 8.64 | 6.62 | 8.64 |
| Patient 9 | 8.64 | 8.64 | 8.62 |
| Patient 10 | 8.66 | 8.66 | 7.69 |
| Patient 11 | 8.56 | 8.56 | 8.55 |

*Table 2 Maximum motion magnitudes for 2 gastric organs (stomach and large bowel) over the 21 phases for all 11 patients that we generated synthetic motion for.*

|  | **Full Body** (mm) | **Stomach** (mm) | Large Bowel (mm) |
|---|---|---|---|
| Patient 1 | 0.04 ± 0.01 | 3.05 ± 0.83 | - |
| Patient 2 | 0.03 ± 0.01 | 2.85 ± 0.9 | - |
| Patient 3 | 0.04 ± 0.02 | 2.22 ± 0.72 | - |
| Patient 4 | 0.01 ± 0.01 | 2.32 ± 0.65 | - |
| Patient 5 | 0.06 ± 0.02 | 3.27 ± 0.96 | - |
| Patient 6 | 0.04 ± 0.02 | 3.24 ± 0.82 | - |
| Patient 7 | 0.06 ± 0.0015 | 0.92 ± 0.03 | 0.77 ± 0.03 |
| Patient 8 | 0.08 ± 0.0006 | 2.57 ± 0.01 | 2.74 ± 0.04 |
| Patient 9 | 0.08 ± 0.0005 | 3.65 ± 0.05 | 2.96 ± 0.01 |
| Patient 10 | 0.07 ± 0.0007 | 3.57 ± 0.03 | 2.78 ± 0.05 |
| Patient 11 | 0.09 ± 0.0006 | 3.69 ± 0.01 | 2.96 ± 0.03 |

*Table 3 Means and standard deviations of motion magnitudes for 2 gastric organs (stomach and large bowel) over the 21 phases for all 11 patients that we generated synthetic motion for.*

| Metric | DIR Algorithm | T1w gaSOS | | CECT | | |
|---|---|---|---|---|---|---|
| | | Patient 1 | Patient 2 | Patient 3 | Patient 4 | Patient 5 |
| | | Stomach & Duo | Stomach & Duo | Stomach | Stomach | Stomach |
| TRE ↓ (mm) | HSOF | 2.14 ± 1.31 | 2.70 ± 1.94 | 3.06 ± 1.62 | 3.33 ± 1.93 | 3.11 ± 2.01 |
| | EVO | 3.02 ± 1.70 | 3.64 ± 2.12 | 3.88 ± 2.30 | 3.33 ± 1.90 | 3.44 ± 2.27 |
| | Elastix | 3.40 ± 2.19 | 2.58 ± 1.45 | 2.48 ± 1.33 | 3.48 ± 2.12 | 3.48 ± 2.41 |
| | Demons | 3.26 ± 1.93 | 3.08 ± 1.94 | 3.27 ± 1.68 | 3.60 ± 2.00 | 3.46 ± 2.05 |
| DSC ↑ | HSOF | 0.81 | 0.78 | 0.99 | 0.79 | 0.77 |
| | EVO | 0.77 | 0.81 | 0.98 | 0.82 | 0.81 |
| | Elastix | 0.68 | 0.77 | 0.98 | 0.76 | 0.71 |
| | Demons | 0.74 | 0.8 | 0.99 | 0.79 | 0.77 |
| HD95 ↓ (mm) | HSOF | 6.00 | 7.00 | 1.00 | 7.48 | 8.31 |
| | EVO | 8.06 | 8.77 | 1.73 | 6.71 | 8.06 |
| | Elastix | 8.06 | 7.35 | 2.24 | 7.18 | 8.66 |
| | Demons | 8.06 | 6.16 | 1.41 | 7.34 | 8.83 |

*Table 4 Segmentation accuracy of various variational registration methods applied to T1WI gaSOS, and CECT. Stomach & Duo: duodenum was included in the segmentation mask of the stomach.*

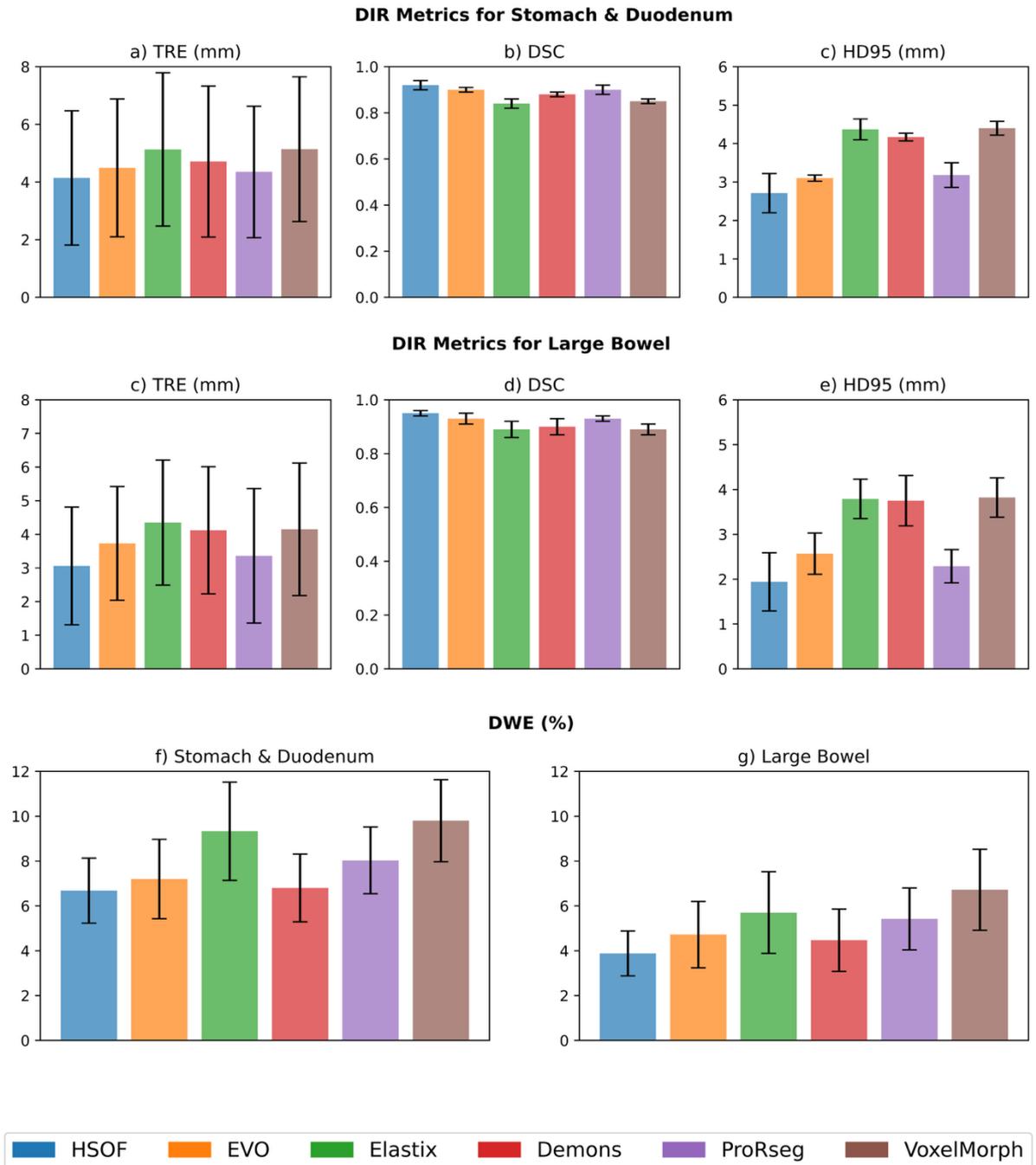

*Figure 4 For five T2w MRI datasets, we report segmentation performance (DSC and HD95), registration accuracy (TRE), and the DWE across various deep learning and variational DIR methods. Motion was applied separately to the stomach, duodenum, and large bowel in each dataset.*

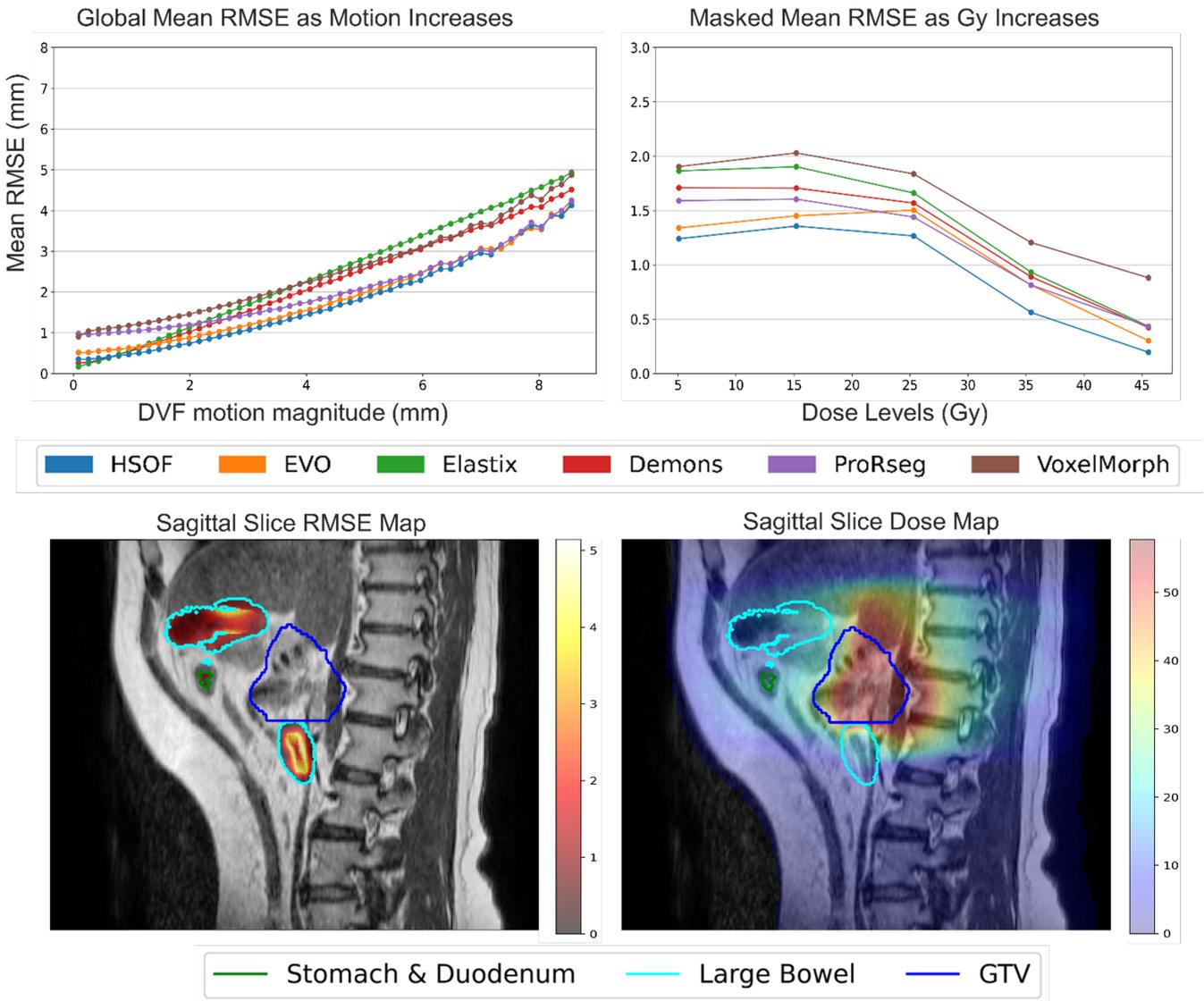

*Figure 5 Patient-Specific granular performance of the different DIR methods on patient A where the Stomach, Duodenum and Large Bowel have a small overlap with the high radiation zone. (a) Global mean RMSE (mm) binned by motion magnitude (b) Stomach, Duodenum and Large Bowel mean RMSE (mm) binned by Gy Radiation Level. (c) Patient Scan visualization with the RMSE overlapped. (d) Patient Scan visualization with the dose map overlapped.*

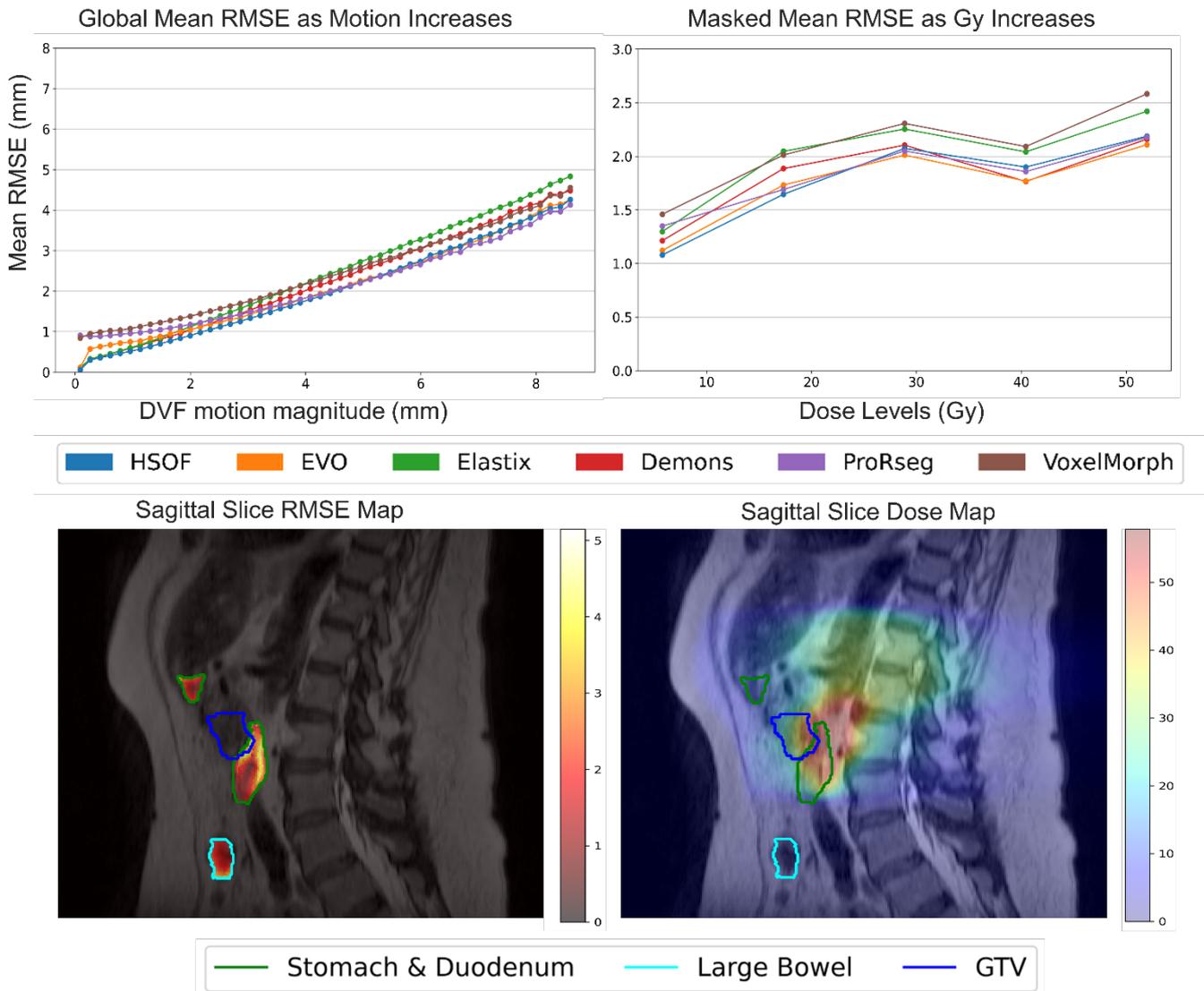

*Figure 6 Patient-Specific granular performance of the different DIR methods on patient B where the Stomach, Duodenum and Large Bowel have a big overlap with the high radiation zone. (a) Global mean RMSE (mm) binned by motion magnitude (b) Stomach, Duodenum and Large Bowel mean RMSE (mm) binned by Gy Radiation Level. (c) Patient Scan visualization with the RMSE overlapped. (d) Patient Scan visualization with the dose map overlapped.*